\documentclass[journal]{journal}
\usepackage{geometry}
\usepackage{cite}
\usepackage{authblk}
\usepackage{multirow}
\usepackage{caption,pdfpages}

\usepackage{amsmath}

\usepackage{atbegshi}
\AtBeginDocument{\AtBeginShipoutNext{\AtBeginShipoutDiscard}}
\begin{document}
%
\title{Loss Function Entropy Regularization for Diverse Decision Boundaries}
%
%
%

\author{Sue Sin Chong}
\affil{Institute for Interdisciplinary Information Sciences, Tsinghua University \\ z18516937829@163.com}
\thanks{Published in International Conferernce on Big Data Analytics 2022 in Guangzhou, China (ICBDA 2022)}

\maketitle

\begin{abstract}
   Is it possible to train several classifiers to perform meaningful crowd-sourcing to produce a better prediction label set without ground-truth annotation? This paper will modify the contrastive learning objectives to automatically train a self-complementing ensemble to produce a state-of-the-art prediction on the CIFAR10 and CIFAR100-20 tasks. This paper will present a straightforward method to modify a single unsupervised classification pipeline to automatically generate an ensemble of neural networks with varied decision boundaries to learn a more extensive feature set of classes. Loss Function Entropy Regularization (LFER) are regularization terms to be added to the pre-training and contrastive learning loss functions. LFER is a gear to modify the entropy state of the output space of unsupervised learning, thereby diversifying the latent representation of decision boundaries of neural networks. Ensemble trained with LFER has higher successful prediction accuracy for samples near decision boundaries. LFER is an adequate gear to perturb decision boundaries and has produced classifiers that beat state-of-the-art at the contrastive learning stage. Experiments show that LFER can produce an ensemble with accuracy comparable to the state-of-the-art yet have varied latent decision boundaries. It allows us to perform meaningful verification for samples near decision boundaries, encouraging the correct classification of near-boundary samples. By compounding the probability of correct prediction of a single sample amongst an ensemble of neural network trained, our method can improve upon a single classifier by denoising and affirming correct feature mappings.
\end{abstract}


%
\IEEEpeerreviewmaketitle

\section{Introduction}

How to make unsupervised algorithms converge and learn different features at will at the same time has always been difficult. Recently, a combination of the representation learning method and end-to-end learning has given promising results in the coarse classification of reasonably large datasets such as CIFAR100 \cite{wvangansbeke2020scan}. SCAN's \cite{wvangansbeke2020scan} three-step process involves two contrastive learning stages and a fine-tuning stage.  

Recently, \cite{JunGuo_AAAI_2020} works on feature selection in unsupervised learning. \cite{pretextInvariant2020}, \cite{pretrain2019Uncurated} works on making the pretext training phase of unsupervised learning more robust. 
The properties of similarity search vectors and application in unsupervised image classification task has also been much studied in  \cite{alex2018spreading}, \cite{spreadLocal2019}. The integration of noise and features into the unsupervised learning pipeline has also proven to be useful \cite{bojanowski2017unsupervised}. \cite{jain2016approximate} \cite{Zhang2015SparseCQ} has studied the possibility of searching in a quantized sparse space representation. \cite{he2019momentum} has studied the momentum of the unsupervised clustering process. Recently, \cite{chen2020simple} introduced a simple contrastive learning framework. Whereas \cite{chuang2020debiased} makes the contrastive learning stage robust. Asano's work \cite{asano2020self} is a fine-tuning technique that has proven to be very effective in the final stages of unsupervised learning. In contrast, Vo's work \cite{vo2019unsupervised} attempts to find new objects or features as an optimization objective. 

Entropy regularization is an important technique in machine learning and has many applications. Several studies \cite{pereyra2017regularizing}, \cite{cuturi2013sinkhorn}, \cite{Liong2015DeepHF} utilized entropy regularization enhance unsupervised learning, entropy regularization can either maximize marginal entropy of bits or speed-up classification. In searching, \cite{subic2017} also utilized entropy regularization on one-hot codes.  Whereas \cite{entropy2015Spring} has studied the mass-spring-damper system without a singular kernel, which might be able to play the role of controller in unsupervised learning. 

\subsection{Related Work}
Out-of-Distribution (OOD) Detection is an increasingly prominent field in machine learning. \cite{pidhorskyi2018generative} is a method for generating a state-of-the-art OOD detector with the adversarial method. \cite{lee2018simple} used a method based on Gaussian to extract features from a trained neural network to train an OOD detector with relatively simple tools. Via \cite{lee2018simple}, one can link the certainty of features learnt to OOD classifier accuracy. 

\subsection{Motivation}
If we can have a check to see whether an unsupervised learner has learned a feature, that would be great. The check serves as a fill-in-the-blank check for an unsupervised classifier. It will allow us to paint a class as a set of more specific and more refined and, at the same time, correct features instead of having to accept a class as a very generic template of features.

Secondly, the ability to separate weakly-classified samples from well-classified samples only by using a similarly-unsupervised-trained neural network or pipeline allows the unsupervised learner to self-check and improve. When it is hard to differentiate, it will be beneficial to be able to have neural networks trained with varied decision boundaries to vote on a particular sample. This motivates us to modify the entropy configuration of output space to train neural networks with varied decision boundaries to help partition the dataset into easy-to-classify and possibly hard-to-classify samples. 

\section{Preliminaries}
We attempt to introduce the possibility of searching in the space of decision boundaries in the unsupervised image classification task. This work attempts to combine the best representation learning, end-to-end learning, and the properties of spreading vector and entropy control to give users a simple yet intuitive way to explore and exploit the entropy dimension in the output space of contrastive learning.
This paper presents a straightforward method for training reasonably good ensembles from a single pipeline with diverse decision boundaries and can learn varied latent representations of the dataset. We think the unsupervised exploration of decision boundaries under varied entropy configurations can improve the unsupervised classification process. We present a framework where we can generate an ensemble that has varied yet complementing decision boundaries simply by changing constants. Our objective is to maximize the number of possibilities in output space entropy distribution of converging neural networks, which allows for a large variety of decision boundaries and, therefore, a highly diversified set of learned latent representations of class features.
LFER is a set of regularization terms added to unsupervised learning objective functions, which will enable unsupervised neural network latent representation exploration. We offer a simple-to-implement yet sensitive gear, allowing for unsupervised learning that exploits differences in the latent representation of decision boundaries to improve accuracy. We have shown that generating an ensemble from an arbitrary deep learning architecture and a dataset (pipeline) is simple and has improved accuracy compared to the best neural network possibly trained from the same pipeline.

The main contributions of our paper are as follows:
\begin{itemize} 
\item We identified a method to exploit entropy in decision boundary formation in the unsupervised classification problem, the LFER method.
\item LFER is simple to implement and use and almost always guarantees to train a neural network with varied decision boundaries from the same dataset from a single architecture. Without LFER, unsupervised learning always converges to the same set of features with the exact decision boundaries,  reproducibly. 
\item LFER method can produce networks with classification accuracy compared to state-of-the-art, but with varied latent decision boundaries. It can also serve as the uniqueness of convergence check on an unsupervised classification pipeline.
\item Ensemble trained with LFER can meaningfully encourage prediction for samples which lie near decision boundaries. 
\item On the CIFAR100-20 task, our ensemble can capture a more extensive super-class feature set.
\item LFER can mine for neural networks, which are better out-of-distribution detectors.
\end{itemize}

In the following section, we will discuss the definition and implications of solving the problem of diversified unsupervised learning with LFER. 

\section{Problem Definition}
Given an unsupervised classification pipeline Pipeline, how can we automatically train an ensemble that can improve prediction accuracy,  where each neural network in the ensemble is independently a reasonable classifier.

\subsection{Applications} LFER is a natural filter for weakly-classified samples in unsupervised learning. With a series of neural networks, which each converge to a different set of correct features, it can immediately identify weakly-classified samples. Secondly, LFER checks for convergence in features learned in unsupervised classification pipelines. When the dataset has multiple ways of matching different features to the same classes, LFER can list the different mappings between features and classes. Finally, it serves as a neighbourhood exploration tool that allows for searching with constants on some arbitrary entropy structure in the output space.

\subsection{Implications} 
When a Pipeline (machine learning architecture + dataset) is sufficient to produce a reasonable classifier by unsupervised learning on a dataset, by using LFER, we can exhaust the feature discovery possibilities of the target Pipeline. Furthermore, suppose there exists any subclass within a superclass. In that case, the LFER will effectively mine for sub-class representations in the original dataset. Finally, there will very likely exist a neural network that converges to an alternative set of features; it serves as a checker for the uniqueness of feature convergence in an unsupervised classification process.

The first section will discuss the LFER method on the Unsupervised Semantic Clustering pipeline. At the same time, the second section will discuss reasoning about entropy regularization and the convergence of neural networks trained with LFER. The third section will discuss the implications of finding complementing neural networks which have learnt different features of the same dataset. The fourth section will discuss the application of combinations of neural networks with different latent decision boundaries. Finally, this paper will conclude with the applications and implications of using LFER as an output space entropy controlling tool. 

\section{Loss Function Entropy Regularization (LFER) }

Loss Function Entropy Regularization (LFER) is a series of entropy regularization terms added to the contrastive learning loss functions. LFER is added to the contrastive learning and pre-training stages of unsupervised classification to encourage different decision boundaries and entropy distribution in the output space, thereby resulting in neural networks with similar accuracy but different latent representations of features. Experiments show that training neural networks with different confusion matrices is impossible without LFER. The implementation of LFER merely requires an additional ten lines of code for the loss function. 

\begin{equation}
\begin{aligned}
(-\lambda_{0}, -\lambda_{1}, +\lambda_{2}, -\lambda_{3}) \\ (& \sum  \Phi_{\eta}^{'c} \log \Phi_{\eta}^{'c}, \sum_{c \in \mathcal{C}}  \Phi_{\eta}^{'c} \log  \Phi_{\eta}^{'c}, \\ & \sum_{c \in \mathcal{C}} \sum_{k \in \mathcal{N}_X} \Phi_{\eta}^{c} \Phi_{\eta}^{k} \log  \Phi_{\eta}^{c}\Phi_{\eta}^{k},  \\  &\sum_{c \in \mathcal{C}}  \sum_{k \in \mathcal{N}_X} \Phi_{\eta}^{'c} \Phi_{\eta}^{k} \log  \Phi_{\eta}^{'c} \Phi_{\eta}^{k} ) \\ & , \space \Phi_{\eta}^{'c} = \frac{1}{|\mathcal{D}|} \sum_{X \in \mathcal{D}}  \Phi_{\eta}^{c} (X))
\end{aligned}
\end{equation}

Entropy terms were previously added in semantic clustering to encourage uniform prediction amongst the contrastive learning stage classes. Our experiment shows that regularization in unsupervised classification objective functions is indispensable for identifying a more extensive set of features, ensemble trained with LFER demonstrates an improvement in the latent representation of features of the neural network.

 Higher-order entropy terms play the role of the controller in controlling the distances between clusters in the output space. Hence it is possible to produce neural networks with varying confusion matrices. Moreover, it is possible to both maximize or minimize between-cluster spaces and the smoothing of the clustering process. 

\subsection{LFER Acting Upon Unsupervised Semantic Clustering}

There are 3 portions to unsupervised classification in the \cite{wvangansbeke2020scan}, SimCLR, Semantic Clustering(SCAN), and Selflabel(SLL).SCAN being contrastive learning stages. Adding entropy terms in the objective functions of contrastive learning stages can improve the diversity of neural networks trained. The regularization term in SimCLR is as follows.

\begin{equation}
\begin{aligned}
& \min_{\theta}d(\Phi_{\theta}(X_i), \Phi_{\theta}(T[X_i])) \\
& - \lambda_{0} \langle \Phi_{\theta}(X_i), \Phi_{\theta}(T[X_i]) \rangle \log \langle \Phi_{\theta}(X_i), \Phi_{\theta}(T[X_i]) \rangle
\end{aligned}
\end{equation}

The regularization term in the SCAN portion of the pipeline is as follows. $\lambda_{2}$ and $\lambda_{3}$ plays the role of a spring term and damper term for between-clusters entropy.

\begin{equation}
\begin{aligned}
\Lambda =   & -\frac{1}{|\mathcal{D}|} \sum_{X \in \mathcal{D}} \sum_{k \in \mathcal{N}_X} \log \langle \Phi_{\eta}(X), \Phi_{\eta}(k) \rangle \\
& +  \lambda_{1} \sum  \Phi_{\eta}^{'c} \log  \Phi_{\eta}^{'c} \\
& - \lambda_{2} \sum_{c \in \mathcal{C}} \sum_{k \in \mathcal{N}_X} \Phi_{\eta}^{c} \Phi_{\eta}^{k} \log  \Phi_{\eta}^{c}\Phi_{\eta}^{k}  \\
& + \lambda_{3} \sum_{c \in \mathcal{C}}  \sum_{k \in \mathcal{N}_X} \Phi_{\eta}^{'c} \Phi_{\eta}^{k} \log  \Phi_{\eta}^{'c} \Phi_{\eta}^{k} \\
& , \space \Phi_{\eta}^{'c} = \frac{1}{|\mathcal{D}|} \sum_{X \in \mathcal{D}}  \Phi_{\eta}^{c} (X)
\end{aligned}
\end{equation}

The scalar terms $(\lambda_i)_{i \geq 0} = f(classes)$ , is a function of number of classes. Below is a table of scalar terms and the classification accuracy of different lambda combinations. The new optimization function presents a control function related to the number of classes to modify the entropy in the input space. Hence we can specify a specific behaviour desired for the output neural network, and then simply by modifying relative values of  $(\lambda_i)_{i \geq 1}$ train a converging network with the desired property.
\begin{table}[h]
\centering
\label{tab:tabela1}
\begin{tabular}{|c|c|c|c|c|c|c|}
\hline
Desc & $\lambda_{0}$ & $\lambda_{1}$ & $\lambda_{2}$ & $\lambda_{3}$ & Scan & SLL  \\ \hline
SCAN & 0 & 5 & 0 & 0 & 0.44 & 0.507       \\
1 & 0 & 5 & 4 & 0 & 0.439 & 0.503 \\
2 & 2 & 5 & 4 & $8$ & 0.4335 & 0.5012 \\
3 & 2 & 5 & 4 & $4$ & 0.4005 & 0.4408 \\
4 & 2 & 5 & $4\sqrt{n}$ & $-8/n$ & 0.456 & 0.4767 \\
\hline
\end{tabular}
\caption{Experiment Results}
\end{table}

Reflecting the sensitivity of spring constant and dampening ratio in a physical spring system, neural networks trained with objective functions with regularization terms that are scalar multiples of each other can also have significantly different decision boundaries. We can also choose to improve the performance of a single classifier in the contrastive learning stage.

\subsubsection{Reasoning About Entropy State}
 By optimizing for entropy exploration, we add the constraint of the number of classes to the output space, forcing different values and structures of classes in the trained neural network. Furthermore, we have introduced an abridged notation to reason about the resulting state of entropy. It will allow for the convenience of solving for deducing the stability of the training process and the properties of the output neural network. The 3-step unsupervised classification process is essentially a cascading function of the entropy exploration and exploitation of the dataset. 

\begin{table}[h]
\centering

\label{tab:tabela1}
\begin{tabular}{|c|c|c|}
\hline
$(\lambda_{i})_{i \geq 1} $ & Value & Notation \\ \hline
$\lambda_{0}$ & 0 & $g(x) = x$ \\
$\lambda_{0}$ & $>0$ & $g(x) = \hat{x}$ \\
\hline
\end{tabular}
\caption{Entropy Regularization State Notation}
\end{table}

Let $g(x)$ be the pretext regularization function. Similarly, we define the following for SCAN. 

\begin{equation}
\begin{aligned}
    \Lambda =   & -\frac{1}{|\mathcal{D}|} \sum_{X \in \mathcal{D}} \sum_{k \in \mathcal{N}_X} \log \langle \Phi_{\eta}(X), \Phi_{\eta}(k) \rangle \\ & + \lambda_{2} x - \lambda_{3}x' + \lambda_{4}x''
\end{aligned}
\end{equation}

Let $\lambda_{2} x - \lambda_{3}x' + \lambda_{4}x'' = h(x) $ be the SCAN regularization function. Then compounding the two regularization functions, the state of entropy resulting from the training process can be expressed as $h(g(x))$. Hence, the contrastive training process maximizes similarity with a second-order differential equation of entropy, offering fine-grained knobs to the desired output entropy state and the decision boundary formation within clusters. LFER serves as a control function for entropy in the contrastive learning stage. 
\subsection{Pseudo Entropy Control Functions}

$(\lambda_{i})_{i \geq 1}$ is a function of number of classes, aka $(\lambda_{i})_{i \geq 1} = f_{i \geq 1} (n) $. Expressing the entropy control function as an inner product between a series of functions of number of classes, we have the following. 

\begin{equation}
\begin{aligned}
    h(g(x)) & = h(n, g(x)) \\
            & = \lambda_{1}g(x) - \lambda_{2}g(x)' + \lambda_{3}g(x)'' \\
            & = f_{1}(n)g(x) - f_{2}(n)g(x)' + f_{3}(n)g(x)'' \\
            & = \langle f_{i \geq 1} (n), g(x) \rangle 
\end{aligned}
\end{equation}

Self-labelling, which attempts to classify noisy neighbours near clusters by minimizing cross-entropy loss correctly, is the only step in the pipeline that optimizes for cross-entropy. LFER makes it possible to systematically produce converging neural networks with different decision boundaries simply by modifying $(f_{i}(n))_{n \geq 1}$. Experiments show that neural networks trained with different objective functions are most confident about a wide variety of prototype images for each super-class. LFER offers a unique edge to the task of CIFAR100-20, as it allows for the mining of more subclass prototypes within a super-class, thereby improving overall classification accuracy.

\subsection{Searching for Constants to Train the Best Performing Neural Network}
In contrastive learning stages, LFER has consistently been able to mine for neural networks that beat state-of-the-art by $1-2\%$ at the end of the contrastive learning stage on the CIFAR10 and CIFAR100-20 classification task. However, the slight edge gained does not persist through the self-labelling stage. LFER can also produce neural networks that train better out-of-distribution classifiers on the same dataset. 

\section{Entropy Perturbation for Decision Boundaries}
LFER is necessary to train neural networks to learn different features. Experiment shows that repeated training on the model without any regularization term results in a similar confusion matrix. The terms force different structures on the neural network decision boundaries, which results in differences in the confusion matrix. 

LFER is a sensitive gear for modifying the entropy environment/configuration in output space. In this section, we will discuss an example ensemble, trained with $\lambda_{3} = \{ 4, 8, 16, 32\} $. To demonstrate that LFER can produce neural networks that learn a different set of features at ease, we introduce the notion of n guess accuracy.  

The top K accuracy is calculated as follows. Given n neural networks, if any n neural network predicts the label correctly, it is calculated as a correct prediction. There is no hierarchy or preference to the set of prediction labels produced by the set of n neural networks. 

\subsection{Self-complementing Ensemble}
The $\lambda_{3}$ term is an effective knob for controlling the fine-grained cluster formation process. It is easy to train a set of neural networks simply by multiplying the $\lambda_{3}$ with a geometric series. By changing $\lambda_{3}$, we are almost certain to find a neural network which has learnt a different set of correct features.

\begin{table}[h]

\begin{center}
\begin{tabular}{|c|c|c|c|c|c|}
    \hline
      Top 2 ACC & Agreement & $\lambda_{0}$ & $\lambda_{1}$ & $\lambda_{2}$ & $\lambda_{3}$ \\
     \hline
     \multirow{ 4}{*}{1}  63.79 & 44.90  & 2 & 5 & 4 & 4 \\
   & & 2 & 5 & 4 & 8\\ \hline
        61.82 & 55.81 & 2 & 5 & 4 & 8 \\
   & & 2 & 5 & 4 & 16\\
    \hline 
   61.8 & 50.59 &2 & 5 & 4 & 16 \\
    & & 2 & 5 & 4 & 32 \\
    \hline
   61.68& 42.64  & 2 & 5 & 4 & 4 \\
    & & 2 & 5 & 4 & 32\\ \hline
\end{tabular}
\end{center}
\label{tab:multicol}
\caption{Top 2 ACC Of a $\lambda_{3}$ Series}
\end{table}

The 2-ensemble agrees on samples in which they are both confident and only disagrees on samples where there is confusion or lies near the decision boundaries. With LFER, we can identify classes and samples where there are second opinions. Therefore it can serve as a filtering method for noisy samples and noisy classes.

\begin{table}[h]

\begin{center}
\begin{tabular}{|c|c|c|c|c|c|}
    \hline
       \multirow{ 5}{*}{1} 3/4 guess ACC & 2-agreement &$\lambda_{0}$ & $\lambda_{1}$ & $\lambda_{2}$ & $\lambda_{3}$ \\
     \hline
     69.35 & 75.86 & 2 & 5 & 4 & 4 \\
    && 2 & 5 & 4 & 8\\
      & & 2 & 5 & 4 & 32\\
    \hline
      67.80 & 83.97  & 2 & 5 & 4 & 4 \\
    && 2 & 5 & 4 & 8\\
      & & 2 & 5 & 4 & 16\\
       \hline
        67.49  & 82.50 & 2 & 5 & 4 & 4 \\
    && 2 & 5 & 4 & 16\\
      & & 2 & 5 & 4 & 32\\
      \hline
        71.92  & NA & 2 & 5 & 4 & 4 \\
    && 2 & 5 & 4 & 8\\
    && 2 & 5 & 4 & 16\\
      & & 2 & 5 & 4 & 32\\ \hline

\end{tabular}
\end{center}
\label{tab:multicol}
\caption{3/4 Guess Of a $\lambda_{3}$ Series}
\end{table}

We sieve for confident samples with majority votes from the ensemble for at least three-quarters of the samples within the ensemble. We can further clamp down on samples that all networks do not agree on, aka samples with very high confusion. Below is an image of the most confident prototypes of an LFER ensemble. This ensemble has learnt the many sub-classes within the 20 superclasses of the CIFAR100 dataset. For instance, in the large carnivores superclass, neural networks in the ensemble have learnt bear, tiger, leopard and lion, respectively, as their most confident prototype for the same superclass. 

\begin{minipage}{\linewidth}
\includegraphics[width = \linewidth]{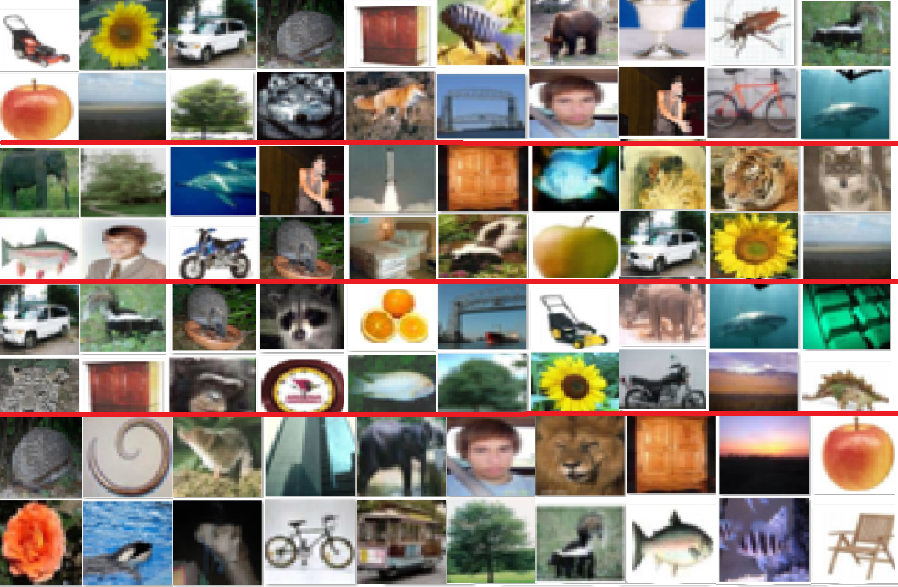}
\captionof{figure}{\textbf{Figure 1. Different Prototype Images of Superclasses}}
\end{minipage}

LFER can mine for a more comprehensive representation of features of classes when there are many sub-classes within a super-class. When sub-classes differ drastically, the advantage will be prominent. 4 neural networks are sufficient for verification checks for weakly-classified samples.

\section{More Entropy Configurations}
 It is also possible to explore the entropy space further with even larger classes of LFER functions. We note that lambdas templates which result in converging neural networks, correspond to spring damper constants which result in spring-damper systems with forced harmonic motion.

\subsection{Extended Stability}
 Experimentation has proven that the following lambda bounds will produce optimization functions that can produce a converging neural network. The number of possible LFER functions which can train converging neural networks in the contrastive learning stage is vast.  

\begin{table}[h]
\centering

\label{tab:tabel3}
\begin{tabular}{|c|c|}
\hline
$(\lambda_{i})_{i \geq 1} $ & Big O Bound  \\ \hline
$\lambda_{0}$ & $O(1)$ \\
$\lambda_{1}$ & $O(1)$ \\
$\lambda_{2}$ & $O(n)$ \\
$\lambda_{3}$ & $O(n\sqrt(n))$ \\
\hline
\end{tabular}
\caption{Big O Bounds for $f(n)$}
\end{table}

\begin{table}[h]
\centering

\label{tab:tablea3}
\begin{tabular}{|c|c|c|c|}
\hline
$\lambda_{0}$ & $\lambda_{1}$ & $\lambda_{2}$ & $\lambda_{3}$  \\ \hline
0 & O(1) & 0 & 0 \\
0 & O(1) & O(1) & 0 \\
O(1) & O(1) & O(1) & 0 \\
O(1) & O(1) & O(1) & O(1) \\
O(1) & O(1) & $O(\sqrt{n})$ & 0 \\
O(1) & O(1) & 0 & $O(n)$ \\
O(1) & O(1) & $O(\sqrt{n})$ & $O(\sqrt{n})$ \\
O(1) & O(1) & $O(1)$ & $O(\frac{1}{n})$  \\
O(1) & O(n) & $O(\frac{1}{n})$ & $O(n\sqrt{n})$ \\
\hline
\end{tabular}
\caption{Stable Templates}
\end{table}

Decision boundaries of a neural network trained are very sensitive to small changes in $(\lambda_{i})_{i \geq 1}$. Therefore, it is reasonable to expect to find neural networks that can identify all other subclass prototypes in the superclass by enumerating the set of $\lambda$s. With LFER, it is possible to have multiple neural networks which are confident about more sub-classes within the superclass to learn the sub-classes within a superclass. 

We observed that the accuracy of a single classifier in CIFAR20 could hardly surpass 0.5 due to limits in the number of classes and the lack of labels. However, having multiple networks learning the 20 classes in CIFAR100-20 makes it possible to have different neural networks learn the features of smaller classes, which results in a better superclass classification accuracy.

\subsection{Combined Ensembles Trained with LFER}
On the CIFAR100 dataset, we have trained 25 Res18Nets for classification into 20 coarse classes. Each of the 25 Res18Nets has been trained with a different lambda constant set. Results in the following table are obtained by combining 2 to 4 of the 25 neural networks. 

\begin{table}[h]
\centering

\label{tab:tabela1}
\begin{tabular}{|c|c|c|c|}
\hline
K & Best ACC & Mean ACC & Median ACC  \\ \hline
1 nn & 50.7 & NA & NA \\
2  & 64.50 & 58.20 & 59.24 \\
3  & 70.57 & 63.70 & 64.70 \\
4  & 73.99 & 68.90 & 69.10 \\
\hline
\end{tabular}
\caption{Top K Performances}
\end{table}
Each classifier is independently an at least $> 0.40$ accuracy classifier. Moreover, $>0.85$ of the classifiers show at least $0.1$ improvement in classification guess. When we compound the LFER of different classes, it is much more likely to make the n guess of the neural networks more robust. Below is a listing of some of the best combinations of neural networks. Some of the best combinations often involve neural networks with non-zero $\lambda_{0}, \lambda_{2}, \lambda_{3}$, demonstrating the indispensability of the LFER in training neural networks with diverse decision boundaries. 
\begin{table}[ht]

\begin{center}
\begin{tabular}{|c|c|c|c|c|}
 \hline
    \multirow{ 4}{*}{1}
   
      Top 2 ACC & $\lambda_{0}$ & $\lambda_{1}$ & $\lambda_{2}$ & $\lambda_{3}$ \\
     \hline
      64.50  & 0 & 5 & 0 & 0 \\
    & 2 & 5 & 4 & 0\\
    \hline
   64.12 & 0 & 5 & 0 & 0 \\
    & 2 & 5 & $1 /4n^2$ & $1/2n$ \\
    \hline
   63.79 & 2 & 5 & 4 & 8 \\
    & 2 & 5 & 4 & 4\\ \hline
\end{tabular}
\end{center}
\label{tab:multicol}
\caption{Top 2 ACC of Combined Ensembles}
\end{table}

\begin{table}[h]

\begin{center}
\begin{tabular}{|c|c|c|c|c|}
    \hline 
     \multirow{ 4}{*}{1} Top 3 ACC & $\lambda_{0}$ & $\lambda_{1}$ & $\lambda_{2}$ & $\lambda_{3}$ \\
     \hline
      70.79  & 0 & 5 & 0 & 0 \\
    & 2 & 5 & 4 & 0\\
    & 2 & 5 & $1/4n$ & $0.5/\sqrt{n}$ \\
    \hline
     70.66 & 0 & 5 & 0 & 0 \\
    & 2 & 5 & 4& 0 \\
    &2 & 5 & 4 & 4 \\
    \hline
     70.60 & 0 & 5 & 0 & 0 \\
    & 2 & 5 & $1/4n$ & $0.5/\sqrt{n}$ \\
     & 2 & 5 & $1 /4n^2$ & $1/2n$ \\ \hline
\end{tabular}
\end{center}
\caption{Top 3 ACC of Combined Ensembles}
\label{tab:multicol}
\end{table}

In the CIFAR100-20 task, there are five subclasses in each superclass; there are multiple ways of mapping sub-classes to a superclass. LFER mines the input dataset for multiple sub-classes within the superclass. By compounding  2 to 3 neural networks whereby each network learns a subclass within a superclass, the n guess predictions will, in expectation, make reasonable guesses which will include each subclass. 

\begin{minipage}{\linewidth}
\includegraphics[width = \linewidth]{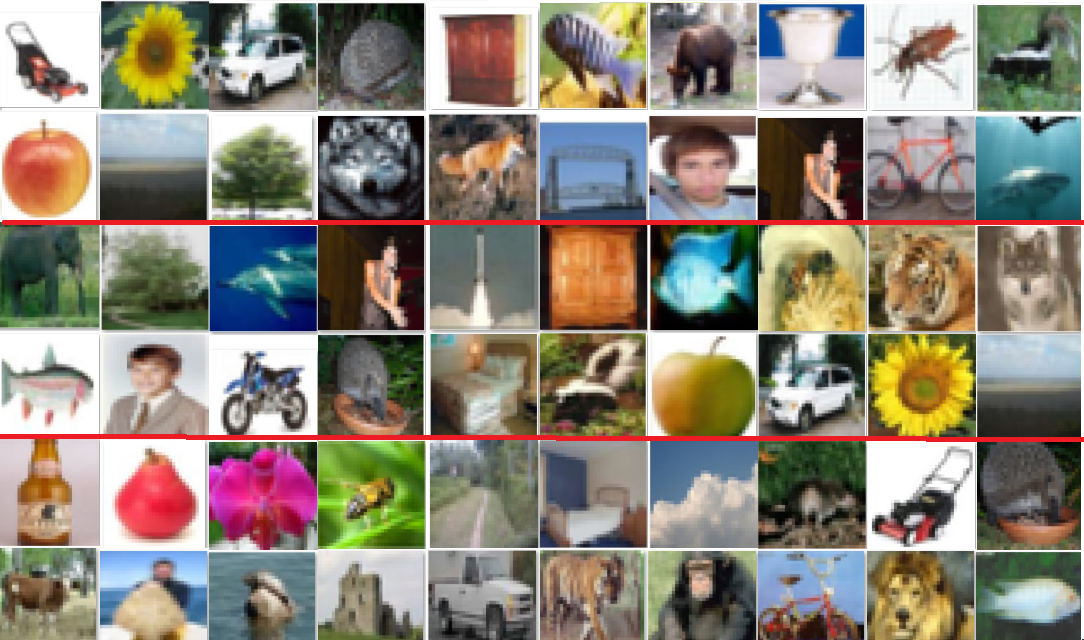}
\captionof{figure}{\textbf{Prototype Images for 2 Guess Combined Ensembles}}
\end{minipage}

Best 3 combinations often have confident prototypes drawn from a larger variety of sub-classes. 
\begin{minipage}{\linewidth}
\includegraphics[width = \linewidth]{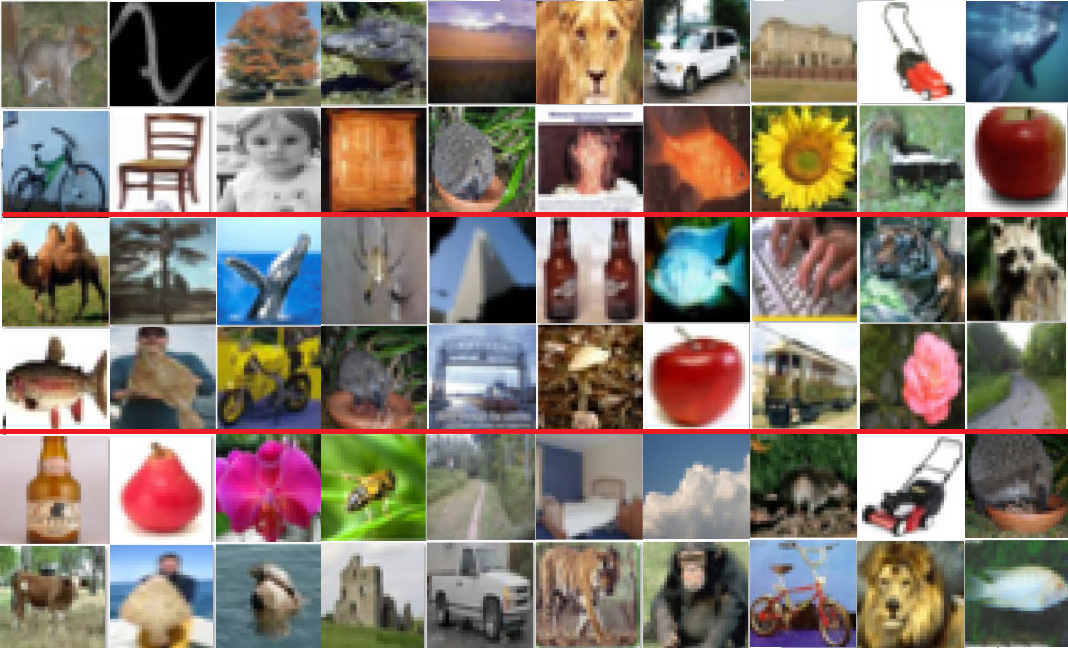}
\captionof{figure}{\textbf{Prototype Images for 3 Guess Combined Ensembles}}
\end{minipage}
While there may be confusion across super-classes, each subclass is only represented once in each classifier confident prototype. Combinations in aggregation learn to be confident about a large set of sub-classes.  

\subsubsection{Learning Sub-classes Within Super-class}
With the LFER, we can mine for different prototypical images within a super-class in the CIFAR100-20 task. Instead of being forced to view a super-class as a set of non-descriptive generic features, LFER can mine for decision boundaries which respect specific subclass features within a super-class. 

\section{Majority Vote}
Training several neural networks and having them vote on classes of a sample can produce state-of-the-art prediction accuracy. N guess accuracy is the upper bound of majority vote accuracy. We can improve upon the accuracy of a single prediction set by compounding the accuracy of several classifiers. Our best combination, which involves 27 different neural networks voting for the label of every single sample, reaches a new state-of-the-art accuracy rate of 0.58. 

\begin{table}[h]

\begin{center}
\begin{tabular}{|c|c|}
    \hline
       Number of Classifiers & Accuracy by Majority Vote  \\  \hline
       State-of-the-Art & 50.7 \\ 
       3 & 55.9 \\
       4 & 56.1 \\
       27 & 58.1 \\ \hline
\end{tabular}
\end{center}
\label{tab:multicol}
\caption{State-of-the-Art Results on CIFAR100-20 by Majority Voting}
\end{table}

Experiment results demonstrate that the fuzziness induced by LFER allows for meaningful verification checks across neural networks trained from the same architecture. We also observed that dividing the classifiers into three tiers according to their accuracy rate and organising voting amongst neural networks drawn from each of the tiers more easily produce an accuracy rate higher than randomly selected neural networks. 

\section{Conclusion}
Neural networks trained with LFER have different latent representations of decision boundaries. The knobs presented by $(\lambda_{i})_{i \geq 1}$ are fine-grained decision boundaries gear for modifying the entropy environment of output space, which influences decision boundary formation. LFER presents the possibility of using simple search techniques on $\lambda$s to automatically mine-train neural ensembles with varied decision boundaries by exploiting the differences in a wide array of latent representations of decision boundaries about a single dataset. As a result, ensembles trained with LFER can often beat the accuracy of the best neural network possible for a particular architecture on a dataset.


%

\ifCLASSOPTIONcaptionsoff
  \newpage
\fi



\bibliographystyle{IEEEtran}
%

\bibliography{References}

\end{document}